\documentclass[runningheads]{llncs}

\title{Structured Active Inference}%
\titlerunning{Structured Active Inference}

\author{Toby St Clere Smithe\orcidID{0000-0002-8317-8722}}
\authorrunning{T. St Clere Smithe}

\institute{VERSES Research Lab \\
\email{toby.smithe@verses.ai}}

\usepackage[utf8]{inputenc}
\usepackage[T1]{fontenc} %

\usepackage{microtype}
\usepackage{float}

\usepackage{amsfonts}
\usepackage{amssymb}
\usepackage{amsthm}
\usepackage{amsmath}
\usepackage{caption}
\usepackage[inline]{enumitem}
\setlist{itemsep=0em, topsep=0em, parsep=0em}
\setlist[enumerate]{label=(\alph*)}
\usepackage{etoolbox}
\usepackage{stmaryrd}
\usepackage{ifthen}
\usepackage{suffix}
\usepackage{verbatim} %

\usepackage{xfrac}

\usepackage{bm}
\usepackage{physics}
\usepackage[safe]{tipa} %
\usepackage{mathtools} %
\usepackage{color}
\usepackage{wrapfig}
\usepackage{tikz}
\usepackage{tikzit}
\usetikzlibrary{cd, babel}

\usepackage[framemethod=tikz]{mdframed}

\newcommand*{\secref}[1]{\S\ref{#1}}

\usepackage[backend=biber,style=numeric-comp,sorting=none,date=short]{biblatex}
\appto{\bibsetup}{\sloppy}

\usepackage[bookmarks=true,pdfusetitle]{hyperref}
\hypersetup{colorlinks,linkcolor=darkblue,citecolor=darkblue,urlcolor=darkblue,breaklinks=true}

\newcommand*{\citep}[1]{\parencite{#1}}
\newcommand*{\citet}[1]{\textcite{#1}}

\usepackage[capitalize]{cleveref} %

\usepackage{graphicx}

\usepackage{pict2e}

\makeatletter
\newcommand{\adjunction}{\@ifstar\named@adjunction\normal@adjunction}
\newcommand{\normal@adjunction}[4]{%
  #1\colon #2%
  \mathrel{\vcenter{%
    \offinterlineskip\m@th
    \ialign{%
      \hfil$##$\hfil\cr
      \longrightharpoonup\cr
      \noalign{\kern-.3ex}
      \smallbot\cr
      \longleftharpoondown\cr
    }%
  }}%
  #3 \noloc #4%
}
\newcommand{\named@adjunction}[4]{%
  #2%
  \mathrel{\vcenter{%
    \offinterlineskip\m@th
    \ialign{%
      \hfil$##$\hfil\cr
      \scriptstyle#1\cr
      \noalign{\kern.1ex}
      \longrightharpoonup\cr
      \noalign{\kern-.3ex}
      \smallbot\cr
      \longleftharpoondown\cr
      \scriptstyle#4\cr
    }%
  }}%
  #3%
}
\newcommand{\longrightharpoonup}{\relbar\joinrel\rightharpoonup}
\newcommand{\longleftharpoondown}{\leftharpoondown\joinrel\relbar}
\newcommand\noloc{%
  \nobreak
  \mspace{6mu plus 1mu}
  {:}
  \nonscript\mkern-\thinmuskip
  \mathpunct{}
  \mspace{2mu}
}
\newcommand{\smallbot}{%
  \begingroup\setlength\unitlength{.15em}%
  \begin{picture}(1,1)
  \roundcap
  \polyline(0,0)(1,0)
  \polyline(0.5,0)(0.5,1)
  \end{picture}%
  \endgroup
}
\makeatother

\let\op=\relax

\def\op{\ensuremath{^{\,\mathrm{op}}}}

\DeclareMathOperator*{\E}{\mathbb{E}}

\renewcommand{\d}{\mathrm{d}}

\newcommand{\Ca}{{\mathcal{C}}}
\newcommand{\Da}{{\mathcal{D}}}

\newcommand{\Rb}{{\mathbb{R}}}

\newcommand{\Vb}{{\mathbb{V}}}

\newcommand{\xto}[2][]{\xrightarrow[#1]{#2}}

\makeatletter
\newcommand{\mathoverlap}[2]{\mathpalette\mathoverlap@{{#1}{#2}}}
\newcommand{\mathoverlap@}[2]{\mathoverlap@@{#1}#2}
\newcommand{\mathoverlap@@}[3]{\ooalign{$\m@th#1#2$\crcr\hidewidth$\m@th#1#3$\hidewidth}}
\makeatother

\newcommand{\kcirc}{\bullet} %

\makeatletter
\providecommand*{\xmapstofill@}{%
  \arrowfill@{\mapstochar\relbar}\relbar\rightarrow
}
\providecommand*{\xmapsto}[2][]{%
  \ext@arrow 0395\xmapstofill@{#1}{#2}%
}

\makeatletter
\def\slashedarrowfill@#1#2#3#4#5{%
  $\m@th\thickmuskip0mu\medmuskip\thickmuskip\thinmuskip\thickmuskip
   \relax#5#1\mkern-7mu%
   \cleaders\hbox{$#5\mkern-2mu#2\mkern-2mu$}\hfill
   \mathclap{#3}\mathclap{#2}%
   \cleaders\hbox{$#5\mkern-2mu#2\mkern-2mu$}\hfill
   \mkern-7mu#4$%
}
\def\rightslashedarrowfill@{%
  \slashedarrowfill@\relbar\relbar\mapstochar\rightarrow}
\newcommand\xslashedrightarrow[2][]{%
  \ext@arrow 0055{\rightslashedarrowfill@}{#1}{#2}}
\makeatother

\theoremstyle{definition}
\newtheorem{defn}{Definition}[section]

\newtheorem*{rmk*}{Remark}

\newtheorem{prop*}{Proposition}

\newtheorem{lemma}[defn]{Lemma}

\newtheorem*{thm*}{Theorem}
\newtheorem*{cor*}{Corollary}

\theoremstyle{remark}

\definecolor{darkblue}{rgb}{0,0,0.7}

\tikzstyle{dot}=[inner sep=0.0mm, outer sep=0.0mm, minimum size=1mm, draw, shape=circle]
\tikzstyle{dot-5mm}=[dot, minimum size=5mm]
\tikzstyle{dot-1cm}=[dot, minimum size=1cm]
\tikzstyle{wcopy}=[dot, fill=white, scale=2.0]
\tikzstyle{bcopy}=[dot, fill=black, scale=2.0]
\tikzstyle{box}=[fill=white, draw=black, shape=rectangle]
\tikzstyle{box-5mm}=[box, minimum size=5mm, shape aspect=1]
\tikzstyle{box-7mm}=[box, minimum size=7mm, shape aspect=1]
\tikzstyle{box-1cm}=[box, minimum size=10mm, shape aspect=1]
\tikzstyle{dia}=[fill=white, draw=black, shape=diamond]
\tikzstyle{dia-5mm}=[dia, minimum size=5mm, shape aspect=1]
\tikzstyle{dia-7mm}=[dia, minimum size=7mm, shape aspect=1]
\tikzstyle{effect}=[regular polygon, regular polygon sides=3, draw]
\tikzstyle{state0}=[regular polygon, regular polygon sides=3, draw, shape border rotate=0]
\tikzstyle{state90}=[regular polygon, regular polygon sides=3, draw, shape border rotate=90]
\tikzstyle{state180}=[regular polygon, regular polygon sides=3, draw, shape border rotate=180]
\tikzstyle{state270}=[regular polygon, regular polygon sides=3, draw, shape border rotate=270]
\tikzstyle{scalar}=[diamond, draw, inner sep=1pt]
\tikzstyle{ground0}=[my ground, draw, inner sep=0pt, minimum width=4.2pt, minimum height=11.2pt, anchor=input, rotate=0]
\tikzstyle{ground90}=[my ground, draw, inner sep=0pt, minimum width=4.2pt, minimum height=11.2pt, anchor=input, rotate=90]

\input{strings.tikzdefs}

\ifexternalizetikz\tikzexternaldisable\fi
\newsavebox\sbground
\savebox\sbground{%
  \begin{tikzpicture}[baseline=0pt]
    \draw (0,-.1ex) to (0,.85ex)
    node[ground IEC,draw,anchor=input,inner sep=0pt,
    minimum width=3.15pt,minimum height=8.4pt,rotate=90] {};
  \end{tikzpicture}%
}

\newsavebox\sbcopy
\savebox\sbcopy{%
  \begin{tikzpicture}[baseline=0pt]
    \node[wcopy,scale=0.7] (a) at (0,3.8pt) {};
    \draw (a) -- +(-90:.21);
    \draw (a) -- +(45:.21);
    \draw (a) -- +(135:.21);
  \end{tikzpicture}}

\ifexternalizetikz\tikzexternalenable\fi

\newsavebox\bsbcopy
\savebox\bsbcopy{%
  \begin{tikzpicture}[baseline=0pt]
    \node[bcopy,scale=0.7] (a) at (0,3.8pt) {};
    \draw (a) -- +(-90:.21);
    \draw (a) -- +(45:.21);
    \draw (a) -- +(135:.21);
  \end{tikzpicture}}

\ifexternalizetikz\tikzexternalenable\fi

\usepackage{mathabx}
\usepackage{quiver}

\def\id{\mathsf{id}}
\def\kto{\rightsquigarrow}

\newcommand{\xkto}[1]{\overset{#1}{\kto}}
\def\xto{\xrightarrow}

\def\cf{\mathfrak{c}}

\def\Agent{\mathbf{Agent}}
\def\Cat{\mathbf{Cat}}
\def\DMoore{\mathsf{DMoore}}
\def\FinStoch{\mathbf{FinStoch}}
\def\Int{\mathbf{Int}}
\def\Gen{\mathsf{Gen}}
\def\Lens{\mathbf{Lens}}
\def\Moore{\mathsf{Moore}}
\def\Poly{\mathbf{Poly}}
\def\Set{\mathbf{Set}}
\def\Sys{\mathsf{Sys}}

\begin{document}

\maketitle

\begin{abstract}
  We introduce \textit{structured active inference}, a large generalization and formalization of active inference using the tools of categorical systems theory.
  We cast generative models formally as systems ``on an interface'', with the latter being a compositional abstraction of the usual notion of Markov blanket; agents are then `controllers' for their generative models, formally dual to them.
  This opens the active inference landscape to new horizons, such as:
  agents with structured interfaces (\textit{e.g.} with `mode-dependence', or that interact with computer APIs);
  agents that can manage other agents;
  and `meta-agents', that use active inference to change their (internal or external) structure.
  With structured interfaces, we also gain structured (`typed') policies, which are amenable to formal verification, an important step towards safe artificial agents.
  Moreover, we can make use of categorical logic to describe express agents' goals as formal predicates, whose satisfaction may be dependent on the interaction context.
  This points towards powerful compositional tools to constrain and control self-organizing ensembles of agents.
\end{abstract}

As active inference develops beyond toy models and early explorations, and becomes a fully-fledged account of the form and function of complex living systems at all scales, the necessity of a compositional organizing framework becomes apparent: without this, the complexity of veridical multi-agent systems in realistic environments is impossible to handle.
To see this necessity, one need only ponder the question, how would you model an agent as it gets on a bicycle, or starts to drive a car?
Even in cases as mundane as these, current presentations of active inference do not yield a canonical answer.
More generally, it is not at all clear how to describe, in general, an agent that can change its `interface'.

This is an argument that we must move beyond the current paradigm of partially observable Markov decision problems (and their continuous-time cousins), which are most suitable for the description of individual agents with unchanging Markov blankets and static environments, and adopt a new paradigm with change at its heart, and that makes the most of the compositional structure of our world.
It is such a paradigm that we propose here: \textit{structured active inference}, in which every agent's generative model has an explicit interface, vastly generalizing the existing notion of Markov blanket.
This new paradigm exploits the new mathematics of categorical systems theory (CST) \parencite{Myers2022Categorical}, which organizes interacting dynamical systems into a coherent compositional framework.

Being based on category theory, change and composition are at the heart of structured active inference: patterns of interaction among systems constitute morphisms of interfaces, as do changes-of-interface; and changes-of-structure are morphisms of systems (over interfaces).
Category theory is moreover the mathematics of structure, meaning that agents' interfaces may exhibit substantially richer structures than merely a fixed pair of sets (of possible actions and observations).
In particular, agents may now exhibit mode-dependence: the available actions may depend on contextual data (encoded in their current state).
Being a mathematical \textit{lingua franca}, category theory allows us to import ideas from computer science, yielding agents that correctly interact with computer APIs, or whose policies may be interpreted as programs and thus type-checked (a step towards formal verification), or that (recursively) manage other agents, or (`corecursively') use active inference to change their own structure.

Additionally, categorical systems theory is itself amenable to categorical logic \parencite{Jacobs1999Categorical}, which explains the structural fundaments of logic and type theory and which may be instantiated in any sufficiently structured context.
In this way, structured active inference permits not only typed policies, but also a rigorous analysis of agents' goals as formal predicates (possibly in a fuzzy, quantitative, temporal logic), whose satisfaction may be judged in relation to their interaction contexts: thanks to its compositionality, the logic of systems is itself `covariant' with their patterns of interaction.
This promises a further important step towards safe artificial agents, as agents' goals may be gated behind the certification (within bounds) of their safety \parencite{Dalrymple2024Safeguarded,Dalrymple2024Guaranteed}, and their policies judged accordingly.
In combination with structurally nested (`manager') agents, it becomes possible to pass formal specifications of high-level goals down to low-level agents which actually implement them.

This extended abstract is an announcement of work in progress.
A full expository paper will follow soon.
Meanwhile, we supply appendices below which sketch the key structures and exemplify the corresponding possibilities.

\begin{credits}
  \subsubsection*{\ackname}
  The author thanks VERSES for financial support during the development of this work.

  \subsubsection*{\discintname}
  The author has no competing interests to declare that are relevant to the content of this article.
\end{credits}

\newpage
\appendix
\section{Categorical systems theory for active inference}

\subsection{Systems indexed by interfaces} \label{sec:moore}

A \textit{systems theory} is a family of categories of systems, covariantly indexed by a category of interfaces.
More formally, this is a (pseudo) functor $\Sys:\Int\to\Cat$, where $\Cat$ is the 2-category of categories, functors and natural transformations.

The category $\Int$ is usually understood to have `interfaces' (of some type) for objects, and ``patterns of interaction'' or ``wiring diagrams'' for morphisms.
Thus, if $p$ and $q$ are two interfaces (objects in $\Int$), then a morphism $p\to q$ encodes ``how to wire a $p$-shaped system into a $q$-shaped system''.

A standard example category of interfaces is the category $\Lens(\Set)$ of `lenses' in $\Set$ (the latter being the category of sets and functions between them).
In this case, the objects of $\Lens(\Set)$ are pairs $(O,I)$ of sets; we think of $O$ as representing the output type of some system, and $I$ as representing its input type. and a morphism (a so-called `lens') $\varphi:(O,I)\to(P,J)$ is a pair of functions $\bigl(\varphi_1:O\to P, \varphi^\sharp:O\times J\to I\bigr)$.
We typically understand the map $\varphi_1$ as representing how an $O$ `output' is mapped to a $P$ output, and the map $\varphi^\sharp$ as representing how an $I$ `input' may be obtained from an $O$ output and a $J$ input.
That is, $\varphi$ encodes how an inner system with output-input interface $(O,I)$ may be `nested' inside an outer system with interface $(P,J)$: the outer outputs are obtained as a transformation of the inner outputs, and the inner inputs are obtained as a transformation of the inner outputs and outer inputs, thereby enabling feedback.

A simple example of a systems theory over $\Lens(\Set)$ is that of deterministic automata, or \textit{Moore machines}, $\Moore:\Lens(\Set)\to\Cat$.
A Moore machine with interface $(O,I)$ is a triple $(S,\vartheta_o,\vartheta^u)$ of a state space $S:\Set$, an output map $\vartheta_o:S\to O$, and an update map $\vartheta^u:S\times I\to S$.
A morphism $f:(S,\vartheta)\to(S',\vartheta')$ of Moore machines over $(O,I)$ is a function $f:S\to S'$ that ``commutes with the dynamics'': \textit{i.e.}, satisfying $f\circ\vartheta^u = {\vartheta'}^u\circ(f\times\id_I)$.
The category $\Moore(O,I)$ is thus constituted by Moore machines with interface $(O,I)$ and their morphisms, which compose simply as functions.

Given a lens $\varphi:(O,I)\to(P,J)$, $\Moore$ must define a functor $\Moore(\varphi):\Moore(O,I)\to\Moore(P,J)$ which `rewires' machines over $(O,I)$ so that they become machines over $(P,J)$.
Note that a Moore machine $(S,\vartheta)$ over $(O,I)$ may equivalently be presented as a lens $(\vartheta_o,\vartheta^u):(S,S)\to(O,I)$, and so $\Moore(\varphi)$ may be defined simply by composing $\varphi$ after $\vartheta$, thus mapping $(S,\vartheta)$ to $(S,\varphi\circ\vartheta)$.
It is easy to check that this maps morphisms of Moore machines over $(O,I)$ to morphisms of Moore machines over $(P,J)$ (and does so functorially); and since composition is functorial, this definition yields a valid pseudo functor.

\subsection{Placing systems side-by-side}

In order to understand morphisms of interfaces as wiring systems together, we need to be able to place systems side-by-side, so that we can make sense of the plural `systems'.
This demands that we can first compose interfaces ``in parallel'', and second that our systems theories are such that we can take a system over one interface and a system over another, and form a system over the corresponding parallel composite interface.
Formally, this means we require $\Int$ to be a monoidal category $(\Int,y,\otimes)$ with parallel composition (`tensor') $\otimes$ and unit (trivial) interface $y$, and that our systems theories are (lax) monoidal with respect to this parallel composition.
The latter requirement means that we need a natural transformation $\Sys(p)\times\Sys(q)\to\Sys(p\otimes q)$ (satisfying some standard coherence conditions).
The laxness of this structure means that this transformation is not invertible: that is, not every system over a composite interface $p\otimes q$ may be decomposable into systems over the component interfaces $p,q$.

$\Lens(\Set)$ is monoidal, with tensor $(O,I)\otimes (A,B) = (O\times A,I\times B)$ and unit $(1,1)$.
$\Moore$ is similarly lax monoidal, with the laxator $\Moore(O,I)\times\Moore(A,B)\to\Moore\bigl((O,I)\otimes(A,B)\bigr)$ defined by mapping $(S,\vartheta),(T,\psi)$ to $(S\times T,\vartheta\otimes\psi)$.

\subsection{Dynamic wiring} \label{sec:internal-hom}

We will see below that it can be useful to define systems with ``dynamic wiring''.
Recalling that `wiring' corresponds to a morphism in a category of interfaces $\Int$, a system that can change some wiring would need an interface type that encapsulates morphisms of interfaces; that is, given interfaces $q,r$ to be wired together, we need an object in $\Int$ that `internalizes' the set of morphisms $\Int(q,r)$.
Such an object is called an \textit{internal hom}, and we will denote it by $[q,r]$.

For $[q,r]$ to behave like an internalization of $\Int(q,r)$, it needs to exhibit a `currying'\footnote{
The \textit{currying} of a function $f:A\times B\to C$ is the function $f^\sharp:A\to[B,C]$ defined by the mapping $a\mapsto f(a,-)$.}
relationship with $\otimes$.
That is, morphisms $p\otimes q\to r$ must be in bijection with morphisms $p\to[q,r]$.
In other words, we need a natural isomorphism $\Int(p\otimes q,r) \cong \Int(p,[q,r])$.
(This makes $[q,-]$ \textit{right adjoint} to $-\otimes q$.)
From this natural isomorphism we can obtain\footnote{
Set $p=[q,r]$ and then transport the identity morphism $\id_{[q,r]}:[q,r]\to[q,r]$ from the right-hand side to the left-hand side.}
`evaluation' morphisms $[q,r]\otimes q\to r$ which say that, given a $[q,r]$-system and a $q$-system, we can wire the former into the latter to obtain an $r$-system.
Thus $[q,r]$-systems dynamically wire $q$-systems into $r$-systems.

We will use this structure to describe agents that manage other agents, generalizing hierarchical (`deep') active inference.

\subsection{Polynomial interfaces}

A lens interface is a fixed pair of sets, much as in classical active inference.
To capture systems whose interfaces may be mode- or context-dependent\footnote{
For example, when I close my eyes, the sense-data that I receive (my `inputs') are different from when I open them.},
we need the possible inputs to a system to be able to depend on the system's state, as revealed by its output.
To encode such dependence, we may use \textit{polynomial functors}, whose morphisms are sometimes known as \textit{dependent lenses}.

A \textit{polynomial} $p$ in one variable, denoted $y$, is an expression of the form $\sum_{i:I} y^{p[i]}$.
We call $I$ the \textit{indexing set} and, for each $i:I$, $p[i]$ is the exponent at $i$.
In high-school algebra, the variable $y$ takes numerical values, the exponents $p[i]$ are all numbers, and the whole expression may be evaluated into a number, thus defining a function.
Here, we allow the variable $y$ and exponents $p[i]$ to be valued in sets, noting that each natural number $n$ may be understood as a finite set $[n]$ with $n$-many elements.
An exponential $B^A$ of sets is understood to denote the set $\Set(A,B)$ of functions $A\to B$, so that each term $y^{p[i]}$ denotes the `hom' functor $X\mapsto X^{p[i]}$, and the sum denotes the disjoint union of sets.
In this way, each polynomial $p$ does indeed denote a functor $\Set\to\Set$.
If we evaluate this functor at the singleton set $1$, we find
\[ p(1) = \sum_{i:I} 1^{p[i]} \cong \sum_{i:I} 1 \cong I \; . \]
For this reason, we will often write a polynomial $p$ as $\sum_{i:p(1)} y^{p[i]}$.

Each polynomial $p$ also corresponds to a (discrete) bundle of sets, given by projecting from the disjoint union of the exponents\footnote{
The elements of $\sum_{i:p(1)}p[i]$ are pairs $(i,x)$ of $i\in p(1)$ and $x\in p[i]$.}
back to the indexing set.
That is, each $p$ corresponds to a function
\begin{align*} \sum_{i:p(1)} p[i] &\to p(1) \\ (i,x) &\mapsto i \; . \end{align*}
Note that, if the exponents do not vary with the index, \textit{e.g.} $q = \sum_{j:J} y^Q$, then the bundle corresponds to a product projection $\sum_{j:J} Q \cong J\times Q \to J$.
Such `non-dependent' polynomials are called \textit{monomials} and can be written as $Jy^Q$.

A morphism $\varphi:p\to q$ of polynomial functors is a natural transformation.
One can show (using the Yoneda Lemma) that these natural transformations correspond to pairs of functions $(\varphi_1,\varphi^\sharp)$ as in the following commutative diagram, where the square on the right is a \textit{pullback} square:
\[\begin{tikzcd}[sep=scriptsize]
	{\sum_{i:p(1)}p[i]} && {\sum_{i:p(1)}q[\varphi_1(i)]} && {\sum_{j:q(1)}q[j]} \\
	\\
	&& {p(1)} && {q(1)}
	\arrow["p"', from=1-1, to=3-3]
	\arrow["{\varphi^\sharp}"', from=1-3, to=1-1]
	\arrow[from=1-3, to=1-5]
	\arrow["{\varphi_1^*q}", from=1-3, to=3-3]
	\arrow["\lrcorner"{anchor=center, pos=0.125}, draw=none, from=1-3, to=3-5]
	\arrow["q", from=1-5, to=3-5]
	\arrow["{\varphi_1}", from=3-3, to=3-5]
\end{tikzcd}\]
Alternatively, $\varphi^\sharp$ may be understood as a $p(1)$-indexed family of functions $\{\varphi^\sharp_i:q[\varphi_1(i)]\to p[i]\}_{i:p(1)}$.
Thus we have a `forwards' function $\varphi_1$ and a $p(1)$-dependent `backwards' function $\varphi^\sharp$.

Polynomial functors and their morphisms constitute a category $\Poly$.
The composition of $\varphi:p\to q$ and $\psi:q\to r$ is given by $(\psi_1\circ\varphi_1, \varphi^\sharp\circ\varphi_1^*\psi^\sharp)$.
Written in indexed form, the backwards components are
\[ r[\psi_1(\varphi_1(i))] \xto{\psi^\sharp_{\varphi_1(i)}} q[\varphi_1(i)] \xto{\varphi^\sharp_i} p[i] \; . \]

We mentioned above that morphisms between polynomial functors may be understood as `dependent' lenses.
To make this perspective clearer, we observe that $\Lens(\Set)$ embeds into $\Poly$.
Each lens object $(O,I)$ corresponds (bijectively) to a monomial $Oy^I$.
Each lens $(O,I)\to(P,J)$ is then exactly a morphism $Oy^I\to Py^J$.
We can therefore think of the exponents of a polynomial as `mode-dependent' input types, and the indexing sets as possible outputs.
Alternatively, we may understand $p(1)$ as the set of ``possible configurations'' that a $p$-shaped system may adopt; and each $p[i]$ is the set of possible incoming signals in configuration $i$, or the ``$i$-context-dependent sensorium''.

\subsection{Tensor and internal hom of polynomials}

$\Poly$ has a `parallel' tensor product $\otimes$ with unit $y$.
$p\otimes q$ is defined as the polynomial $\sum_{(i,j):p(1)\times q(1)} y^{p[i]\times q[j]}$; likewise, one takes products on morphisms.
It is easy to verify that $y\otimes p \cong p \cong p\otimes y$, and that this product is symmetric, $p\otimes q \cong q\otimes p$.

This tensor structure $(y,\otimes)$ has a corresponding internal hom $[-,=]$.
On objects, the internal hom may be defined by
\[ [p,q] = \sum_{\varphi:p\to q} y^{\sum_{i:p(1)} q[\varphi_1(i)]} \; . \]
(Given morphisms $\varphi:p'\to p$ and $\psi:q\to q'$, we leave the definition of $[\varphi,\psi]:[p,q]\to[p',q']$ as an exercise for the reader.)

We therefore think of the configurations of $[p,q]$ as encoding wirings from $p$ to $q$; and, for each such configuration $\varphi$, the inputs $[p,q][\varphi]$ are the inputs demanded by $\varphi^\sharp$ --- thus, the data required to pass inputs to any $p$-system wired by $\varphi$ into $q$.

For much more on polynomial functors, we refer the reader to the wonderful book by \textcite{Niu2023Polynomial}.

\subsection{Generative models as stochastic Moore machines} \label{sec:gen}

Above (\secref{sec:moore}), we introduced the theory of Moore machines over $\Lens(\Set)$, where we saw that a Moore machine on $(O,I)$ is a choice of state space $S$ and lens $(S,S)\to(O,I)$.
We have just seen that $\Lens(\Set)$ is the monomial subcategory of $\Poly$, so a (non-dependent) Moore machine over $(O,I)$ may alternatively be defined as a $\Poly$ morphism $Sy^S\to Oy^I$ for some set $S$.
This yields a first generalization: a \textit{dependent} Moore machine with polynomial interface $p$ is a pair $(S,\vartheta)$ where $S$ is a set and $\vartheta$ a morphism $Sy^S\to p$ in $\Poly$.
By similar reasoning to that in \secref{sec:moore}, this yields a systems theory $\DMoore:\Poly\to\Cat$.

The reader may have noticed a similarity between the data of a Moore machine $\vartheta:Sy^S\to Oy^A$ --- consisting of an output map (or `likelihood') $\vartheta_o:S\to O$ and an update map (or ``transition function'') $\vartheta^u:S\times A\to S$ --- and a (discrete-time) generative model with state space $S$, observations $O$, and actions $A$.
There are two key differences: a generative model is a \textit{stochastic} transition system; and it is usually understood as being equipped with a `prior' distribution (a stochastic initial condition) on the state space $S$.

Let us address the first difference first, defining a systems theory $\Gen:\Poly\to\Cat$ of stochastic dependent Moore machines, which we take as a mode-dependent generalization of classical generative models.
To keep the exposition simple, we will restrict ourselves to discrete probability distributions, which may be defined over any set.
If $X$ is any set, we will write $\Da X$ to denote the set of such distributions over $X$.
That is,
\[ \Da X = \left\{ p: X\to [0,1] \;\middle|\; \sum_{x:X} p(x) = 1 \right\} \; . \]
For the sum over $X$ to be well-defined, $p\in \Da X$ may only be supported on a countable subset of $X$.

Given any function $f:X\to Y$, we have a function $\Da f:\Da X\to \Da Y$ defined by `pushforward': if $p\in\Da X$, then $\Da(f)(p)$ is the distribution mapping $y\in Y$ to $\sum_{x\in f^{-1}\{y\}} p(x)$.
(The pushforward distribution $\Da(f)(p)$ is sometimes also denoted by $f_*p$.)
This makes $\Da$ into a functor $\Set\to\Set$.

A conditional probability distribution $q(y|x)$ then corresponds to a function $X\to\Da Y$; note that, by currying, this is equivalent to a stochastic matrix $X\times Y\to[0,1]$.
If $q:X\to\Da Y$ and $r:Y\to\Da Z$ are two stochastic matrices, then they may be composed by matrix multiplication to yield a conditional distribution $r\kcirc q:X\to\Da Z$ defined by $(r\kcirc q)(z|x) = \sum_{y:Y} r(z|y)\,q(y|z)$.
Thus conditional distributions form the morphisms of a category, $\FinStoch$.
These morphisms are also known as \textit{stochastic maps} or \textit{channels}.
We will denote a channel $X\to \Da Y$ by $X\kto Y$
The identity channel $X\kto X$ is defined by mapping $x$ to the Dirac delta distribution at $x$; this mapping is sometimes denoted $\eta_X$.
Every function $f:X\to Y$ yields a channel $\delta_f:X\kto Y$ defined by $X\xto{f}Y\xto{\eta_Y}\Da Y$.

A (discrete-time, discrete-space) Markov chain thus corresponds to a channel $X\kto X$, for some state space $X$.
We will use this correspondence to define the systems theory $\Gen$.

Given a polynomial $p$, we define the objects of $\Gen(p)$ to be triples $(S,\vartheta_o,\vartheta^u)$ where $S$ is a set, $\vartheta_o$ is a function $S\to p(1)$ and $\vartheta^u$ is a channel $\sum_{s:S} p[\vartheta_o(s)] \kto S$.
We will call these \textit{generative models} or \textit{stochastic dependent Moore machines} with interface $p$.
A morphism $f:(S,\vartheta)\to(S',\vartheta')$ in $\Gen(p)$ is a channel $f:S\kto S'$ such that $f\kcirc\vartheta^u = {\vartheta'}^u\kcirc(f\times\id)$.
Morphisms of generative models compose by channel composition.

Given a dependent lens $\varphi:p\to q$, we obtain a functor $\Gen(\varphi):\Gen(p)\to\Gen(q)$ by post-composition, almost exactly as for Moore machines.
Thus, $\Gen(\varphi)$ maps a model $(S,\vartheta_o,\vartheta^u)$ over $p$ to the model $(S,\varphi_1\circ\vartheta_o,\vartheta^u\kcirc\delta_{\vartheta_o^*\varphi^\sharp})$.
More explicitly, the update channel is here given by the composite
\[ \sum_{s:S} q[\varphi_1(\vartheta_o(s))] \xkto{\delta_{\vartheta_o^*\varphi^\sharp}} \sum_{s:S} p[\vartheta_o(s)] \xkto{\vartheta^u} S \]
whose first component is the deterministic channel induced by the function $\vartheta_o^*\varphi^\sharp$.
It is easy to check that this definition preserves morphisms of models and thus yields a valid pseudo functor $\Gen:\Poly\to\Cat$.

It will be important in the sequel that $\Gen$ is monoidal, so that we can place generative models in parallel.
Thus, we need a family of functors $\lambda_{p,q}:\Gen(p)\times\Gen(q) \to \Gen(p\otimes q)$ natural in the polynomials $p,q$.
These are defined as follows.

First, note that there is a family of functions $\Da X\times\Da Y\to\Da(X\times Y)$, natural in the sets $X$ and $Y$, given by mapping $\alpha\in\Da X$ and $\beta\in\Da Y$ to the joint distribution $\alpha\otimes\beta$ whose independent marginals are $\alpha$ and $\beta$.
That is, $\alpha\otimes\beta$ is defined by mapping $(x,y)\in X\times Y$ to the probability $\alpha(x)\,\beta(y)$.
(Note that this $\otimes$ is not the same as the tensor on $\Poly$; it is rather the tensor on $\FinStoch$.)

Next, we define $\lambda_{p,q}$ using these functions.
If $\vartheta=(S,\vartheta_o,\vartheta^u)$ is a model over $p$ and $\chi=(T,\chi_o,\chi^u)$ is a model over $q$, then $\lambda_{p,q}(\vartheta,\chi)$ is defined by $(S\times T, \vartheta_o\times\chi_o, \vartheta^u\otimes\chi^u)$.
In turn, $\vartheta^u\otimes\chi^u$ is defined by the function
\[ \sum_{(s,t):S\times T} p[\vartheta_o(s)]\times q[\chi_o(t)] \xto{\vartheta^u\times\chi^u} \Da S\times \Da T \xto{\otimes} \Da(S\times T) \; . \]
If $f:(S,\vartheta_o,\vartheta^u)\to(S',\vartheta'_o,{\vartheta'}^u)$ and $g:(T,\chi_o,\chi^u)\to(T',\chi'_o,{\chi'}^u)$ are morphisms in $\Gen(p)$ and $\Gen(q)$ respectively, then $\lambda_{p,q}(f,g)$ is the morphism $\lambda_{p,q}(\vartheta,\chi)\to\lambda_{p,q}(\vartheta',\chi')$ defined by the function
\[ S\times T \xto{f\times g} \Da S'\times \Da T' \xto{\otimes} \Da(S'\times T') \; . \]
One may then check that this makes $\lambda_{p,q}$ into a functor, and that this definition is natural in $p$ and $q$ (\textit{i.e.}, that it respects morphisms on both interfaces).

\begin{rmk*}
  To check that $\Gen$ really does correspond to classical (discrete time and space) generative models means checking how it behaves over monomials $Oy^A$.
  The objects of $\Gen(Oy^A)$ are triples $(S:\Set, \vartheta_o:S\to O, \vartheta^u:S\times A\kto S)$.
  Thus they are \textit{almost}, but not quite, exactly classical generative models: the output maps are still deterministic by this definition.

  The reason for this is technical: in general, it is so that the pullback set $\sum_{s:S} p[\vartheta_o(s)]$ is well defined.
  In the monomial case, there is no problem having stochasticity in $\vartheta_o$, because the defining expression should evaluate to $S\times A$ in any case.
  But it is less simple to make sense of the expression $\sum_{s:S} p[\vartheta_o(s)]$ when $\vartheta_o$ is a stochastic map (although there are ways to do so).

  Nonetheless, this is not in the end a problem, because if we want a stochastic `likelihood', we can fold it into the update channel.
  That is, we can let the state space be $S\times p(1)$, the output map be the projection $S\times p(1)\to p(1)$, and the update channel be defined by $\sum_{(s,i):S\times p(1)} p[i] \xkto{(\vartheta^u,\vartheta_l)} S\times p(1)$, where $\vartheta^u$ is a `transition' channel $\sum_{(s,i):S\times p(1)} p[i]\kto S$ and $\vartheta_l$ is a `likelihood' channel $S\kto p(1)$, so that $(\vartheta^u,\vartheta_l)$ maps $(s,i,x)$ to $\vartheta^u(s,i,x)\otimes\vartheta_l(s)$.

  Alternatively, we could make an alternative definition of $\Gen$ so that a model over $p$ consists of a quintuple $(S,O,\vartheta_l,\vartheta_o,\vartheta^u)$ where $S$ and $O$ are sets, $\vartheta_l$ is a likelihood channel $S\kto O$, $\vartheta_o$ is an output function $O\to p(1)$, and $\vartheta^u$ is an update channel $\sum_{o:O} p[\vartheta_o(o)]\kto S$.
  We made the definition we did because it keeps the amount of data to a minimum and maintains a close analogy with deterministic Moore machines.
\end{rmk*}

We end this section by noting that priors over states are simply additional data attached to a generative model.
Thus we can define a variant of $\Gen$, denoted $\Gen_*$, where the objects of each $\Gen_*(p)$ are be tuples $(S,\vartheta,\pi)$ where $(S,\vartheta)$ is an object of $\Gen(p)$ and $\pi:1\kto S$ is a distribution over $S$.
We may similarly extend the definition of morphism so that a morphism $f:(S,\vartheta,\pi)\to(S',\vartheta',\pi')$ in $\Gen_*(p)$ is a morphism $f:(S,\vartheta)\to(S',\vartheta')$ in $\Gen(p)$ satisfying the additional condition that $\pi' = \delta_f\kcirc\pi$.
With this definition, there is a canonical indexed functor $\Gen_*\Rightarrow\Gen$ that simply forgets the data of the priors.
We will see below (\secref{sec:logic}) that this is a basic example of a logical structure over $\Gen$.

\subsection{Agents' control systems are dual to their generative models} \label{sec:duality}

The preceding section introduced structured generative models, but an active inference agent is more than just a generative model: it is a model along with a system for `controlling' the model, supplying inputs (actions) that minimize expected free energy, in an attempt to minimize the divergence between how the world is inferred to be (from the observations) and how the world is expected to be (from the transition model).

In active inference, an important distinction is made between the \textit{generative model}, which forms part of an agent, and the \textit{generative process}, which may be quite different from the generative model, but which is understood as the ``ground truth'' generator of the agent's observations (and which is the `true' recipient of the agent's actions).
The generative model is used by the agent to predict how the generative process (\textit{i.e.}, how the `environment') will respond to its actions.

In order to make such a prediction --- in order to `unroll' the generative model in time --- the agent must supply it with actions, which are chosen by the agent's control system.
Formally, this means that the generative model along with the control system must form a closed system (a system on the trivial interface $y$), because closed systems are those that can be evolved autonomously, without external input.

If the generative model has interface $p$, then this means that the control system should have interface $[p,y]$, because (as we saw in \secref{sec:internal-hom}) there is a canonical morphism $[p,y]\otimes p\to y$.
Thus any $p$-system (any $p$-generative model) may be canonically coupled to a $[p,y]$-system (a $p$-controller) to produce a closed system (with interface $y$), which may be unrolled.

Let us verify that this abstract reasoning captures the structure of a classical active inference agent with interface $Oy^A$.
We have already seen that its generative model comprises a state space $S$, a stochastic transition function $S\times A\kto S$, and a `likelihood' $S\to O$.
The dual\footnote{
Mathematically, many dualities are obtained by ``homming into a unit object'' \parencite{Jacobs2017Recipe}, which is precisely the situation we have here.}
of $Oy^A$ is $[Oy^A,y] = A^Oy^O$.
A Moore machine on this interface comprises a state space $T$, an output map $T\to A^O$, and an update map $T\times O\to T$.
By currying, we can equivalently write the output map as $T\times O\to A$.
If we let the state space $T$ be $\Da S$, then our two maps finally have the types $\Da S\times O\to A$ and $\Da S\times O\to \Da S$.
The latter is the type signature of Bayesian inversion \parencite{Braithwaite2023Compositional}, so this map may be understood as performing state inference (`perception').
The former may then be understood as performing action selection, and we may call it the \textit{policy}; in active inference, this is achieved by the minimization of expected free energy.

By inspection, we can therefore see that the pair of a generative model over $Oy^A$ and a Moore machine over $[Oy^A,y]$ captures all the data of a classsical active inference agent.
This licenses the following important definition.

\begin{defn}
  An \textit{agent} with polynomial interface $p$ and state space $S$ consists of a \textit{generative model} over $p$ with state space $S$, along with a \textit{controller} for the generative model: a Moore machine over $[p,y]$ with state space $\Da S$.
  The output map of the controller is the agent's \textit{policy}, performing action selection; the update map of the controller performs \textit{inference}.
\end{defn}

We emphasize again that the agent's interface $p$ generalizes the classical notion of the Markov blanket of its generative model.

\subsection{Animated categories: generative $\Poly$} \label{sec:gen-poly}

The preceding exposition has discussed individual agents, and justified the constructions using `shallow' generative models with interface $Oy^A$.
But active inference purports to model agents in complex contexts, interacting with other agents; and even a single agent is often supposed to have some `deep' or hierarchical structure \parencite{Friston2017Deep}.

We can incorporate such situations easily into the structured active inference framework, by noticing that $p \cong [y, p]$.
Thus, we can equivalently define a `simple' agent with interface $p$ as a model over $[y,p]$ coupled to a controller over $[p,y]$.
If an agent is embedded into a more complex context, we can model this by replacing the trivial interface $y$ with an interface $q$ describing the ``outer boundary'' of the interaction context.
Thus a `complex' agent with interface $p$ in context $q$ may be defined as a model over $[q,p]$ along with a controller over $[p,q]$.

This situation may be rendered fully compositional using the monoidal structure of $\Gen$ and $\Moore$, formalized in the framework of \textit{animated categories} \parencite{Smithe2023Animating}.
The key observation here is that the internal hom structure $[-,=]$ induces an ``internal composition'' operation in the category of interfaces $\Poly$: there are morphisms $[p,q]\otimes[q,r]\to[p,r]$ which act to ``compose along $q$'', and these are natural in $p,q,r$.
This is to say that $\Poly$ is enriched in itself.

In general, whenever we have a category $\Ca$ enriched in a category $\Vb$, and a monoidal functor $F:\Vb\to\Vb'$, we can ``change the enrichment of $\Ca$ along $F$'', to produce a category $F_*\Ca$ enriched in $\Vb'$.
Here, we have a category $\Poly$ enriched in $\Poly$, and a (pseudo) functor $\Gen:\Poly\to\Cat$.
Thus we can produce a category $\Gen_*\Poly$ enriched in $\Cat$ --- that is, a \textit{bicategory}.
We call $\Gen_*\Poly$ \textit{generative $\Poly$}.
(This is \textit{animation}: change of enrichment along a systems theory.)

The objects of $\Gen_*\Poly$ are polynomials, and the hom-category from $p$ to $q$ is $\Gen([p,q])$.
Thus a 1-morphism $p\to q$ is a generative model on $[p,q]$ and a 2-morphism is a corresponding morphism of systems.
If we have a model $p\to q$ and a model $q\to r$, we can compose them using the internal composition structure.
We will see below (\secref{sec:deep}) that this is a precise generalization of what happens in deep/hierarchical active inference that allows for richer situations like ``agents that manage agents''.

\section{Exemplification}

\subsection{APIs as polynomial interfaces}

Polynomial functors are closely related to trees \parencite{Kock2008Polynomial} and tree-structured data, like typed syntax \parencite{Arkor2020Algebraic}: the allowed inputs to a formal operation depend on the operation in question, and its output may be passed as an input to a subsequent operation, so long as the types match.
Thus, a formal grammar corresponds to a polynomial functor $p$, and the set of $p$-terms with inputs in a set $X$ is the set $p(X)$; an interpretation of the syntax in a set $Y$ is then given by a function $p(X)\to Y$.

In this way, if we have a computer application programming interface (API) or domain-specific language (DSL), we can encode its interaction pattern as a polynomial functor $p$; and if we wish to construct an artificial agent that interacts with the corresponding computer system, we can equip it with an interface derived from $p$.

As a simple example, we might consider a `calculator' language with two binary operations $\{+,\times\}$, one unary operation $\{-\}$, and $\Rb$-many constants.
This would correspond to the polynomial functor $p = \{+,\times\}y^2 + \{-\}y + \Rb$.
A machine that takes an expression with inputs in $X$ and outputs a real number could then have the interface $\Rb y^{pX}$; and a model of such a machine with 

We note that this interface is, in the end, monomial.
But if we had a more complex language with different possible output types, we could encode these as a family of polynomials $\{p_j\}$, yielding a machine with interface $\sum_j O_j y^{p_jX_j}$, where $O_j$ is the output type for $p_j$.

Finally, let us note that this short section only scratches the surface of the the connection between polynomial interfaces and typed languages.

\subsection{Typed policies, structured spaces: actions as morphisms} \label{sec:policy}

In classical active inference, a policy is a sequence of actions.
Because all actions are valid in all states (as necessitated by the monomial formalism), any sequence of actions is a valid policy.
In structured active inference, only certain actions are valid at each state, so only certain sequences constitute valid policies.
Following the link between polynomial functors and typed languages sketched in the preceding section, one could say that, in structured active inference, policies are \textit{typed}: they are sequences of \textit{composable} actions.

In some situations, this result can be strengthened: every category $\Ca$ yields a polynomial $\cf=\sum_{x:\Ca_0} y^{\sum_{y:\Ca_0}\Ca(x,y)}$ whose configurations are the objects of $\Ca$ and whose actions at each object $x$ are the morphisms out of $x$.
We can think of a category as an abstract structured space: morphisms out of $x$ correspond to paths away from $x$.
Alternatively, we can think of a category as a (non-deterministic\footnote{
If we work with categories enriched in weighted sets, meaning each morphism is equipped with a weight, then we can think of such a category as representing a stochastic automaton; or as having paths equipped with a length.})
automaton, whose states are the objects, and whose transitions out of each state $x$ are the morphisms out of $x$.
In such a setting, a policy really is a composable sequence of morphisms.

We emphasize that this is a powerful strengthening of the classical notion of policy.

On the practical side, this strengthening allows us to apply active inference to manipulate structured contexts in a principled way.
Any such structured context may usually be expressed in a categorical way, and using this homoiconicity, we may attach a structured active inference agent to it.
In this way, structured agents may manipulate their own (or other agents') structure, thereby answering the question posed at the opening of this extended abstract.
We will consider this ability in more detail below (\secref{sec:meta}).

On the ethical side, the strengthened notion of policy constitutes a step towards safer artificial agents, first because policies may be type-checked, and second because categories often come with an ``internal logic'', in which propositions may be formulated, and against which policies may be verified.
We will sketch more in this area below (\secref{sec:logic}).

\subsection{Hierarchical agents: systems that manage systems} \label{sec:deep}

In \secref{sec:gen-poly}, we indicated that structured active inference extends naturally to `nested' systems of agents, including to the special case of individual agents with a `deep' or hierarchical structure, by extending the interface of a generative model from $p\cong[y,p]$ to $[q,p]$ for some polynomial $q$.
Here, we sketch how this plays out.

The basic idea is to see $q$ as encoding the interface(s) of systems that may be nested within $p$, just as a dependent lens $q\to p$ encodes the wiring of $q$ into $p$.
The polynomial $q$ might itself be composed as the tensor of a number of other polynomials, such as $q=q_1\otimes q_2\otimes q_3$.
Then a system in $\Gen_*\Poly(q_1\otimes q_2\otimes q_3,p)$ could be understood as one that builds a high-level agent out of three lower-level ones.
We could think of such a system as a `manager' for the $\{q_i\}$ systems, bringing them together to form a `collective' with interface $p$.
Then, if we had three lower-level systems $\{y\to q_i\}$, we could tensor them together and compose them with the manager $q_1\otimes q_2\otimes q_3\to p$ to instantiate the collective $y\to p$.

Of course, an active inference agent is composed of a system and its dual (\secref{sec:duality}), so to achieve a compositional description of nested agents, we need to incorporate both parts of the structure.
This is easily done.

\begin{defn}
  Let the bicategory $\Agent$ be the `diagonal' sub-bicategory of $\Gen_*\Poly\times(\DMoore_*\Poly)\op$ whose objects are pairs $(p,p)$ of two copies of a polynomial $p$; we will write these objects simply as $p$.
  Additionally, restrict the state spaces of the 1-cells to be pairs $(S,\Da S)$ where $S$ is a set.
  Thus a 1-morphism $q\to p$ is a pair of a model in $\Gen_*\Poly(q,p)$ and a corresponding controller in $\DMoore_*\Poly(p,q)$.
\end{defn}

Let us consider then a `manager' agent $By^C\otimes Dy^E \to Oy^A$; recall that $By^C\otimes Dy^E \cong (B\times D)y^{C\times E}$.
Such an agent consists of:
\begin{enumerate}[label=(\alph*)]
\item a \textit{state space} $S$;
\item three output maps (`\textit{likelihoods}'):
  \begin{enumerate}[label=(\roman*)]
  \item $S\times B\times D \to O$, predicting a high-level observation in $O$, on the basis of the state and the low-level predictions/observations in $B$ and $D$;
  \item $S\times B\times D\times A \to C$, returning a low-level action in $C$ (for the left low-level agent), on the basis of the state, the low-level observations, and the high-level action in $A$;
  \item $S\times B\times D\times A \to E$, returning a low-level action in $E$ (for the right low-level agent), on the basis of the state, the low-level observations, and the high-level action in $A$;
  \end{enumerate}
\item a \textit{transition channel} $S\times B\times D\times A\kto S$, updating the state, given low-level observations and high-level action;
\item \textit{generative processes} for both low-level agents' observations, on the basis of the state and the high-level observation;
  \begin{enumerate}[label=(\roman*)]
  \item $\Da S\times O\to B$, for the left low-level agent;
  \item $\Da S\times O\to D$, for the right low-level agent;
  \end{enumerate}
\item a high-level \textit{policy}, $\Da S\times O\times C\times E\to A$, possibly depending on all of: the state, the high-level observation, and the low-level actions;
\item an \textit{inference process}, $\Da S\times O\times C\times E\to \Da S$, inferring the current state given a prior, a high-level observation, and low-level actions (which we take to be observed).
\end{enumerate}

\hfill\newline
As we have seen above (\secref{sec:duality}), an agent with interface $By^C$ is constituted by
\begin{enumerate}[label=(\alph*)]
\item a state space $T$,
\item an output map $T\to B$ (yielding the low-level agent's prediction of the high-level agent's generative process),
\item a transition channel $T\times C\kto T$,
\item a policy $\Da T\times B\kto C$, and
\item an inference process $\Da T\times B\kto \Da T$.
\end{enumerate}
A similar structure is of course obtained for the other low-level agent (on $Cy^E$).
Then the composition of the low-level agents to the manager agent proceeds by passing:
\begin{enumerate}[label=(\alph*)]
\item the manager's low-level action outputs to the low-level agents' transition models;
\item the low-level agents' predicted observations to the manager's transition model;
\item the manager's generated observations to the low-level agent's policies and inference processes; and
\item the low-level agents' actions (selected by their policies) to the high-level agent's policy and inference process.
\end{enumerate}
Put together, this composite agent has the interface $Oy^A$; the internal composition is hidden within this ``generalized Markov blanket''.
\newline

We emphasize that all of this compositional structure is obtained from the rules of the $\Agent$ construction; in principle, it is fully automatable.
Moreover, it extends gracefully to arbitrary levels of nesting, with complex internal structure.

\subsubsection{Deep active inference is a special case.}

We end this section by sketching how the $\Agent$ construction captures deep active inference, in the sense of \textcite{Friston2017Deep}.

A deep active inference agent has a hierarchy of generative models: the outputs (the `predictions') at one level act on the level below, where they are taken as a `control' or `policy' input.
(It may also be the case that each level also produces predictions of other observations, but these are effectively extraneous, in that they have no direct effect on the evolution of the model.)

This pattern is easily captured by a sequence of 1-cells
\[ y\to y^{A_1} \cdots \to y^{A_i}\to y^{A_{i+1}} \cdots \]
in $\Agent$, where the increasing index $i$ corresponds to increasingly high levels in the hierarchy.
We leave it to the reader to write out the details and check the correspondence.
(One may simply recapitulate the example above, taking $B=D=E=O=1$ and letting $C,E$ correspond to two levels $A_i,A_{i+1}$ of the action hierarchy.)

\subsection{Meta-agents: changing internal and external structure} \label{sec:meta}

In \secref{sec:policy}, we explained that categorical structures can be encoded in polynomial interfaces.
Because structured active inference collects agents, and their structure, into categories, this makes it able to act on itself.

Crucially, the models $\Gen(p)$ over each interface $p$ form a category, whose morphisms represent changes of internal structure.
And the morphisms between interfaces (of course) represent changes of interface.

These two types of change can be bundled together by applying the \textit{Grothendieck construction} to $\Gen$.
This yields a category $\int\Gen$ whose objects are triples $(p,S,\vartheta)$  where $p$ is an interface in $\Poly$ and $(S,\vartheta)$ is a model in $\Gen(p)$.
A morphism $(p,S,\vartheta)\to(p',S',\vartheta')$ in $\int\Gen$ is then a pair of a dependent lens $\varphi:p\to p'$ and a morphism of systems $\Gen(\varphi)(S,\vartheta)\to(S',\vartheta')$.
Thus the morphisms of $\int\Gen$ are morphisms of interfaces along with compatible morphisms of systems.

One could then consider an agent whose interface combines the polynomial induced by $\int\Gen$ with observations on ``the current interface'', as in
\[ \sum_{(p,S,\vartheta):\int\Gen} p(1)\,y^{\sum_{(p',S',\vartheta')} \int\Gen\bigl((p,S,\vartheta),(p',S',\vartheta')\bigr)} \; . \]
Such an agent would predict both a structure for an agent along with the observations that that agent should expect, and its policy would be a sequence of changes-of-structure (which include standard operations like model expansion and model reduction).
By minimizing free energy, one might imagine that such an agent would seek a structure that best explains its observations: it would be a \textit{structure-learning agent}.
Planning (policy search) for such an agent would involve exploring different potential structures, which we can interpret as \textit{explanations} for the observed data.

This approach can be extended to hierarchical agents, turning $\Agent$ into a double category\footnote{
The horizontal morphisms would be agents; the vertical morphisms would be changes-of-interface; and the squares would be changes-of-structure compatible with the corresponding changes-of-interface.}.
But we will not expand on this here, except to note that this brings structured active inference even closer to other developments in categorical cybernetics \parencite{Capucci2022Foundations,Capucci2023Diegetic,Smithe2021Cyber}.
We leave this confluence to future work.

\subsection{The logic of systems: goals, constraints, and safety} \label{sec:logic}

Let us end this tour of structured active inference with a note on the logic of systems.
We indicated in \secref{sec:gen} that agents' state priors can be understood as a basic example of a logical structure.
This is due to the (indexed) forgetful functor $U:\Gen_*\to\Gen$.
All that is required to bring logic into a categorical situation is the ability to define a notion of \textit{fibration} in that setting; and indeed the setting of categorical systems theory is sufficiently rich to do so (because systems theories collect into a 2-category).

The forgetful functor $U$ is a simple instance of a fibration of systems theories.
In a logical context, one thinks of the fibre $U_\vartheta$ over an object $\vartheta$ as a category of `predicates' over $\vartheta$.
The reindexing operation entailed by the fibration may be interpreted as substitution of terms inside the formal expressions of these predicates.
When the situation is sufficiently rich, these substitution functors have left and right adjoints which correspond to existential and universal quantification in these predicates.

We will not elaborate the details of the categorical logic of systems further here, referring the interested or intrepid reader instead to \textcite{Jacobs1999Categorical} (and the task of translating that work to the systems-theoretic setting).
But we will note that it is of course possible to produce more interesting fibrations of predicates than $U$, with more expressive capability; and we will be able to use these to specify not only goals for systems, but also constraints on their behaviour (their policies).
Moreover, because these logics are indexed by the systems' patterns of interaction, we will be able to judge the extent to which such constraints are satisfied in different interaction contexts\footnote{
This is formally much like the situation with fixed points: the fixed points of a composite dynamical system may be computed from the fixed points of the composite systems, plus knowledge of their pattern of interaction \parencite{Spivak2015steady}}.
We expect that these abilities will be useful for distributing shared goals amongst collectives of agents, and crucial in the development of verifiably safe (`safeguarded' \parencite{Dalrymple2024Safeguarded}) AI systems.

\end{document}